\documentclass{article}

\PassOptionsToPackage{numbers, compress}{natbib}

\usepackage[final]{neurips_2020}

\usepackage[utf8]{inputenc} 
\usepackage[T1]{fontenc}    
\usepackage{hyperref}       
\usepackage{url}            
\usepackage{booktabs}       
\usepackage{amsfonts}       
\usepackage{nicefrac}       
\usepackage{microtype}      
\usepackage{graphicx}
\usepackage[caption = false]{subfig}

\title{Copyspace: Where to Write on Images?}

\author{%
  Jessica M. Lundin \\
  Salesforce\\
  \texttt{jlundin@salesforce.com} \\
   \And
   Michael Sollami \\
   Salesforce \\
   \texttt{msollami@salesforce.com} \\
   \AND
   Brian Lonsdorf \\
   Salesforce \\
   \And
   Alan Ross \\
   Salesforce \\
   \And
   Owen Schoppe \\
   Salesforce \\
   \And
   David Woodward \\
   Salesforce \\
   \And
   Sönke Rohde \\
   Salesforce \\
}

\begin{document}

\maketitle

\begin{abstract}
The placement of text over an image is an important part of producing high-quality visual designs. Automating this work by determining appropriate position, orientation, and style for textual elements requires understanding the contents of the background image.  We refer to the search for aesthetic parameters of text rendered over images as ``copyspace detection'', noting that this task is distinct from foreground-background separation. We have developed solutions using one and two stage object detection methodologies trained on an expertly labeled data. This workshop will examine such algorithms for copyspace detection and demonstrate their application in generative design models and pipelines such as Einstein Designer~\cite{rohde_2020}. 
\end{abstract}

\section{Introduction}
The primary application areas for copyspace detection includes generation of email banners, hero-pages, and call-to-actions, most often performed manually, with notable advances in generation \cite{vempati2019enabling,luban_2018,Hua_2018}. The process of graphic asset development is time-consuming, requiring designers to curate and manipulate media and vector graphics, build up layer stacks, and finally place and format text, all while balancing style, brand consistency and tone in the design.

\section{Data and Methods}

\begin{figure*}[ht!]
  \subfloat[\label{genworkflow}]{
      \includegraphics[trim=0 0 0 0, clip, width=0.27\textwidth]{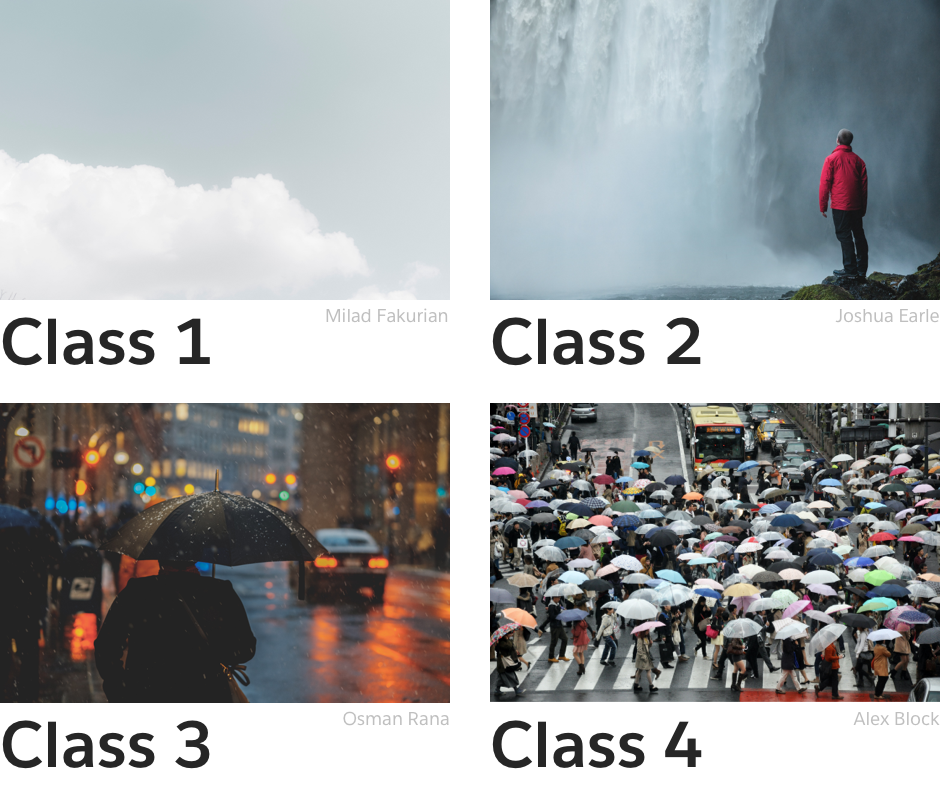}} 
\hspace{\fill}
  \subfloat[\label{synthetic_images}]{
      \includegraphics[trim=0 -60 0 0, clip, width=0.35\textwidth]{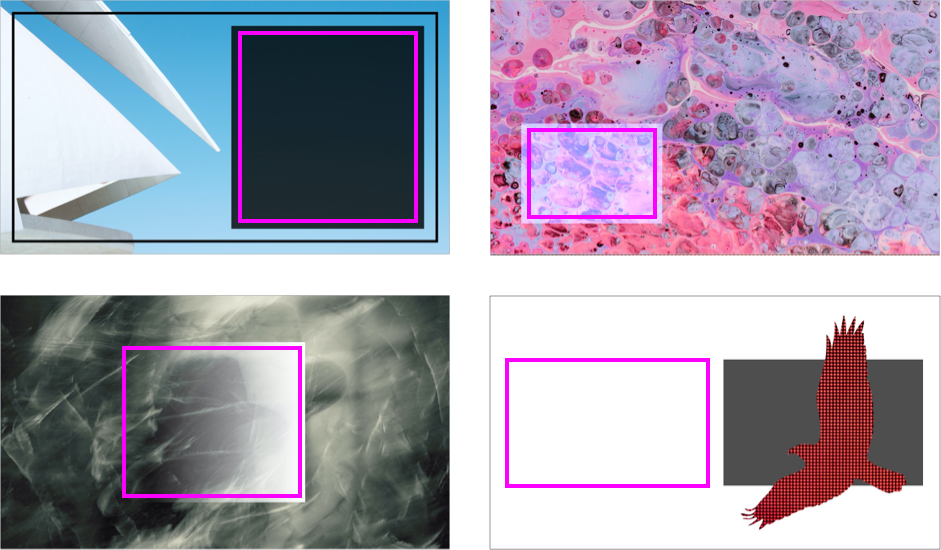}} 
\hspace{\fill}
  \subfloat[\label{mt-simtask}]{
      \includegraphics[trim=0 -160 0 -80, clip, width=0.31\textwidth]{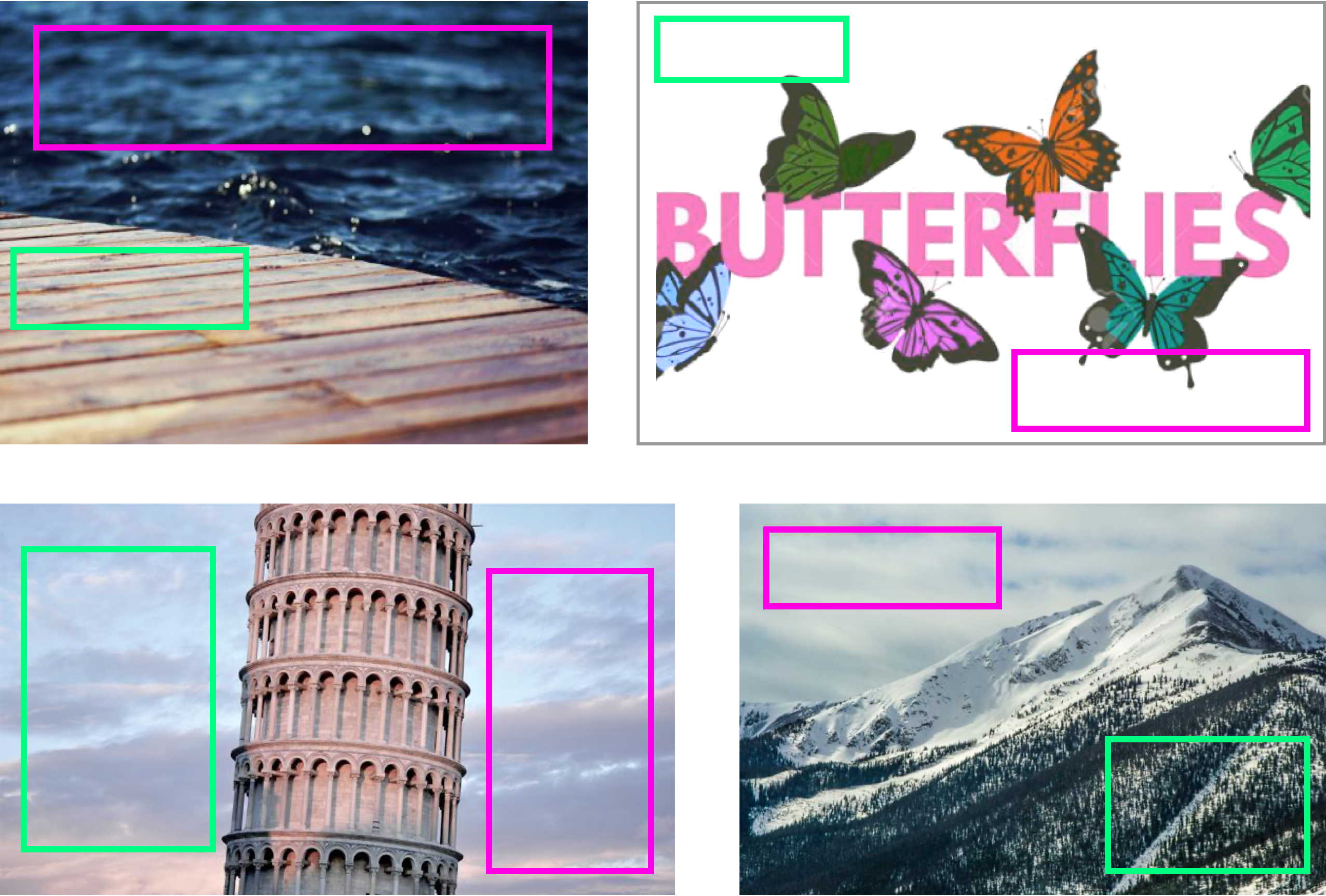}}\\ 
\subfloat[\label{genworkflowcoco}]{
      \includegraphics[trim=0 -50 0 0, clip, width=0.38\textwidth]{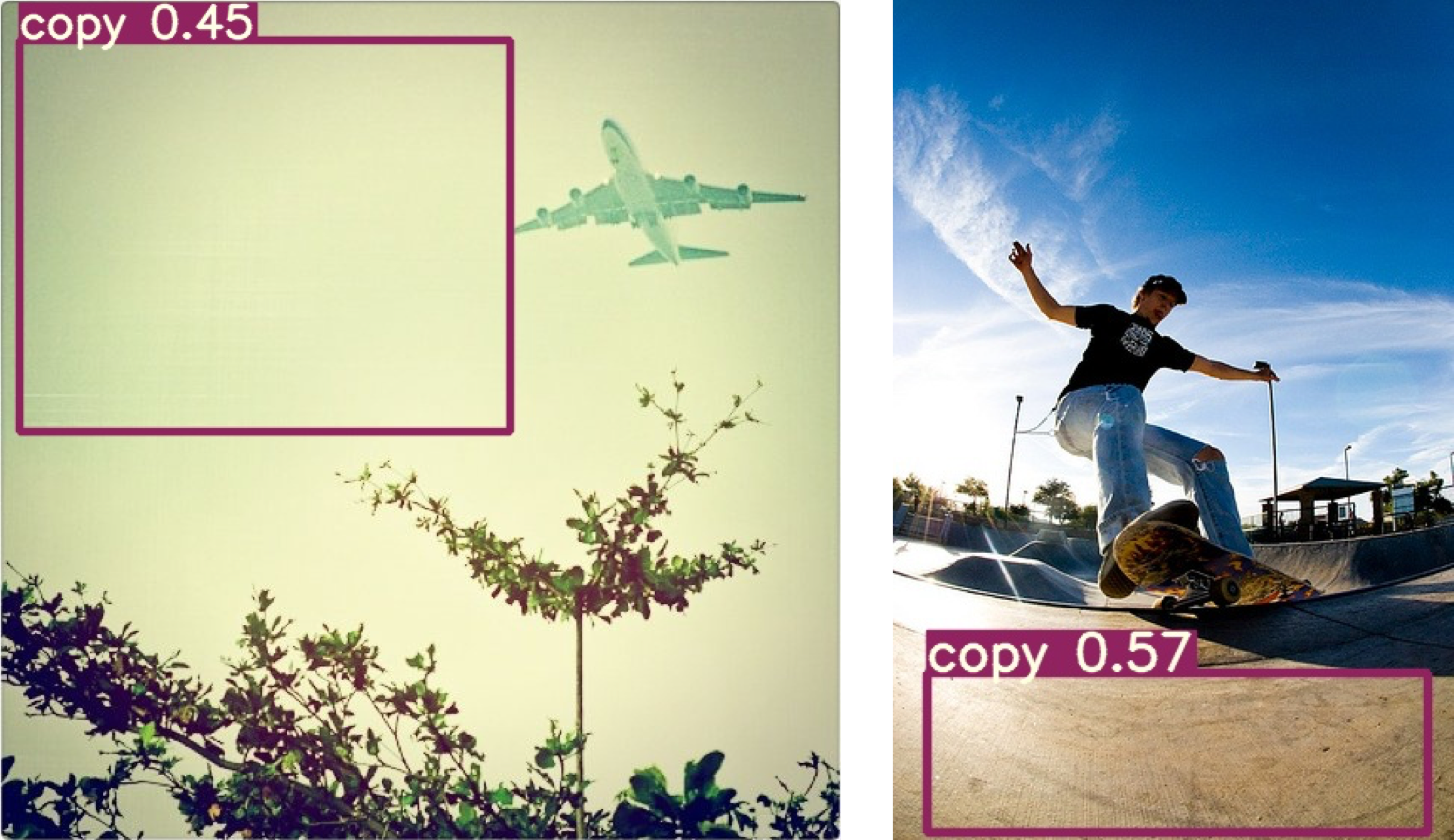}}
\hspace{\fill}
  \subfloat[\label{pyramidprocess}]{
      \includegraphics[trim=0 0 0 0, clip, width=0.59\textwidth]{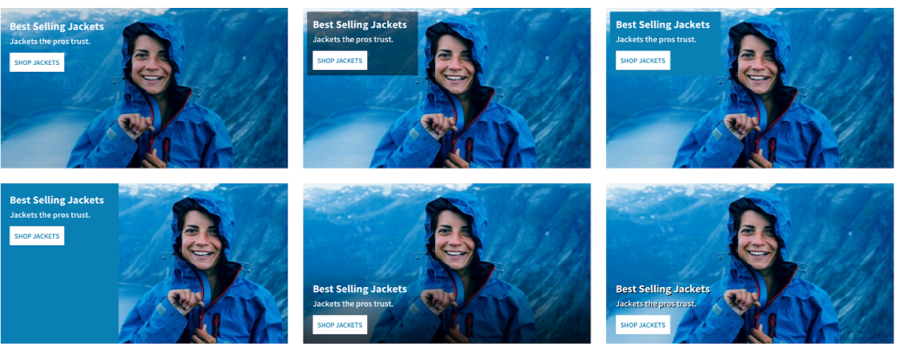}}
\caption{Ground truth and predicted regions are rendered green and magenta respectively. (a) Class 1-4 images indicate increasing difficulty; (b) Human annotated and synthetic labels; (c) Good predictions are sometimes disjoint from annotations; (d) Examples of copyspace detection applied to Coco data set~\cite{lin2015microsoft};
(e) Varying the copyspace algorithm parameters yields multiple ad generations for a single image;
}
\label{fig:results}
\end{figure*}

A set of $20\,000$ license-free images \cite{unsplash} were labeled by a team of experts. High-level design principles and compositional rules from graphic design theory were explicitly encoded within the labels.  Figure \ref{fig:results} shows how we further divide images into four categories of ascending complexity of design.  Additional label synthesis was performed using randomly decorated polygons to augment our training data.

We explore the copyspace problem utilizing frameworks for object detection  \citep{DBLP:journals/corr/RenHG015,DBLP:journals/corr/RedmonDGF15,glenn_jocher_2020_3983579}.  The Yolov5 Github repository is cited in lieu of corresponding publication in arxiv.org, a controversy we will not delve into beyond providing more model inter-comparison results.  

\section{Results and Discussion}

Table \ref{table:1} shows results of a copyspace detection sample intercomparison, where among this set of models regression-based Yolo models generally show higher mAP and IoU performance with fewer parameters.  Among ways copyspace is distinct from object detection is that there is not a single concrete copy space from which to draw rectangular bounds.  When analyzing typical metrics for copyspace including IoU and mAP, it must be taken into account that a reasonable candidate copyspace might not be in the limited set of annotations. 

Table \ref{table:2} shows results for 4 classes of image complexity.  Because the data set is highly imbalanced toward class 1, lower complexity images, we see the mAP results are preferentially biased.  Figure \ref{fig:results} shows inference on Coco images \cite{lin2015microsoft}.  

In this limited exploration of copyspace detection we find favorable initial results using object-detection frameworks.  Machine learning approaches can supplement this nuanced and lower-level task in the designer workflow, and allow focus on higher-skilled tasks.  Copyspace detection is a component of a generative system, and further refinements to this task will directly improve complexity and variation of candidate design generations.

\begin{table}
\begin{tabular}{ c c c c c c c  }
 \hline\hline
     & \textbf{Input Size} & \textbf{Parameters} & \textbf{Layers} & \textbf{mAP @ 0.5} & \textbf{mAP @ .5:.95} & \textbf{IoU}\\
 \hline\hline
 \textit{Yolo v4}     & 416$\times $416 & 2.7e7 & 162 & 20.3 &  -    & 85.1\\
 \textit{Faster RCNN} & 640$\times $640 & 6.4e7 & 290 & 26.4 & 16.3 & 82.1\\
 \textit{Yolo v5s}    & 640$\times $640 & 7.3e6 & 191 & 30.1 & 23.3 & \textbf{88}\\
 \textit{Yolo v5l}    & 640$\times $640 & 4.6e7 & 335 & 32 & 24.4 & 73.9\\
 \textit{Yolo v5x}    & 640$\times $640 & 8.8e7 & 407 & \textbf{34.2} & \textbf{27.2} & 64.4\\
 \hline\hline\\
\end{tabular}
\caption{Comparison of top-performing copyspace detectors.}
\label{table:1}
\end{table}


\begin{table}
\begin{tabular}{ c c c c }
 \hline\hline
     & \textbf{N} &  \textbf{mAP @ 0.5}  & \textbf{mAP @ .5:.95} \\
 \hline\hline
 \textit{Class 1}     & 1.69e3 &   76.2 & 73.5\\
 \textit{Class 2}    & 80  & 34.3 & 19.9 \\
 \textit{Class 3}    & 69  & 17.3 & 11.2   \\
 \textit{Class 4} & 73  & 9 & 4.3\\
 \hline\hline\\
\end{tabular}
\caption{Copyspace class results of image classes for Yolov5x.}
\label{table:2}
\end{table}

\section{Ethics}
Machine learning is a powerful tool that can abstract away tasks of a worker, in this case potentially automating a step in a designer's workflow in asset generation including art, advertisements, etc.  In building out potential tools it is imperative to examine the potential societal impacts.  In medicine there is a saying that a physician should practice at the top of their license.  Abstracting away low-level tasks allows time to be shifted to higher-level tasks.  In this case, the designer may focus more effort on the copy language and font properties that achieve an overarching effect.  While it is not possible to say what the future will yield, there are numerous examples, e.g. ATM machines, where automation provided increased efficiency and jobs.

{\small
\bibliographystyle{ieee_fullname}
\bibliography{copybib}

\begin{thebibliography}{1}\itemsep=-1pt

\bibitem{luban_2018}
Alibaba Cloud.
\newblock Alibaba luban: Ai-based graphic design tool, Dec 2018.

\bibitem{Hua_2018}
Xian-Sheng Hua.
\newblock Challenges and practices of large scale visual intelligence in the
  real-world.
\newblock In {\em Proceedings of the 26th ACM International Conference on
  Multimedia}, page 364. Association for Computing Machinery, 2018.

\bibitem{glenn_jocher_2020_3983579}
Glenn Jocher, Alex Stoken, Jirka Borovec, NanoCode012, ChristopherSTAN, Liu
  Changyu, Laughing, Adam Hogan, lorenzomammana, tkianai, yxNONG, AlexWang1900,
  Laurentiu Diaconu, Marc, wanghaoyang0106, ml5ah, Doug, Hatovix, Jake
  Poznanski, Lijun Yu, changyu98, Prashant Rai, Russ Ferriday, Trevor Sullivan,
  Wang Xinyu, YuriRibeiro, Eduard~Reñé Claramunt, hopesala, pritul dave, and
  yzchen.
\newblock ultralytics/yolov5: v3.0, Aug 2020.
\newblock \url{https://doi.org/10.5281/zenodo.3983579}.

\bibitem{lin2015microsoft}
Tsung-Yi Lin, Michael Maire, Serge Belongie, Lubomir Bourdev, Ross Girshick,
  James Hays, Pietro Perona, Deva Ramanan, C.~Lawrence Zitnick, and Piotr
  Dollár.
\newblock Microsoft coco: Common objects in context, 2015.

\bibitem{DBLP:journals/corr/RedmonDGF15}
Joseph Redmon, Santosh~Kumar Divvala, Ross~B. Girshick, and Ali Farhadi.
\newblock You only look once: Unified, real-time object detection.
\newblock {\em CoRR}, abs/1506.02640, 2015.

\bibitem{DBLP:journals/corr/RenHG015}
Shaoqing Ren, Kaiming He, Ross~B. Girshick, and Jian Sun.
\newblock Faster {R-CNN:} towards real-time object detection with region
  proposal networks.
\newblock {\em CoRR}, abs/1506.01497, 2015.

\bibitem{rohde_2020}
Sönke Rohde.
\newblock Einstein designer — ai-powered, personalized design at scale, Jun
  2020.

\bibitem{unsplash}
Unsplash.
\newblock The internet’s source of freely-usable images, 2020.

\bibitem{vempati2019enabling}
Sreekanth Vempati, Korah~T Malayil, Sruthi V, and Sandeep R.
\newblock Enabling hyper-personalisation: Automated ad creative generation and
  ranking for fashion e-commerce, 2019.

\end{thebibliography}
}

\end{document}